\documentclass[12pt,a4paper]{article}

\usepackage{url}
\usepackage{amsmath}
\usepackage{array,multirow,graphicx}
\usepackage{color}
\usepackage{anyfontsize}
\usepackage{textcomp}
\usepackage{placeins}
\usepackage{fancyvrb}
\usepackage[left=2cm, right=2cm, top=2cm, bottom=2cm]{geometry}

\title{ModelFactory: A Matlab/Octave based toolbox to create human body models}
\author{Manish Sreenivasa\thanks{manishs@uow.edu.au, University of Wollongong, Australia} and Monika Harant\thanks{Optimization, Robotics \& Biomechanics, Heidelberg University, Germany}}
\date{ver. 13.07.2018}
\begin{document}
\maketitle

\begin{abstract}
\noindent Model-based analysis of movements can help better understand human motor control. Here, the models represent the human body as an articulated multi-body system that reflects the characteristics of the human being studied.

\noindent We present an open-source toolbox that allows for the creation of human models with easy-to-setup, customizable configurations. The toolbox scripts are written in Matlab/Octave and provide a command-based interface as well as a graphical interface to construct, visualize and export models. Built-in software modules provide functionalities such as automatic scaling of models based on subject height and weight, custom scaling of segment lengths, mass and inertia, addition of body landmarks, and addition of motion capture markers. Users can set up custom definitions of joints, segments and other body properties using the many included examples as templates. In addition to the human, any number of objects (e.g. exoskeletons, orthoses, prostheses, boxes) can be added to the modeling environment.  

\noindent The ModelFactory toolbox is published as open-source software under the permissive zLib license. The toolbox fulfills an important function by making it easier to create human models, and should be of interest to human movement researchers.

{\bf Please cite this work as: M Sreenivasa, M Harant - arXiv preprint arXiv 1804.03407, 2018, DOI: 10.5281/zenodo.1137656}
\end{abstract}

\section{Background} 
Multi-body models can represent the human body segments and the relative movement between them. Coupled with mechanics-based methods, such models can be a valuable asset in the analysis of human movements. Here, we refer to models that describe the limbs as rigid-segments connected via idealized joints. By utilizing mechanics-based methods (e.g. \cite{Felis2016,FeatherstoneBook}), such models may be used to analyze recorded movements, simulate novel movements, and compute internal body parameters such as joint torques, that may not be easily measured (e.g. \cite{Felis2015,Sreenivasa2016,Sreenivasa2017,Millard2017,Harant2017}). 

An important prerequisite is to create models that can accurately match the body proportions and inertial properties of the human subjects. Additionally, models may only be as detailed as necessary in order to answer specific questions. For example, for motions that occur predominantly in the sagittal plane it may be sufficient to model body segment properties only in that plane (e.g \cite{Sreenivasa2017,Millard2017,Harant2017,Geyer2010}). On the other hand, for more complex movements such as balancing \cite{Sreenivasa2012}, or atypical bone geometry \cite{Sreenivasa2016}, it may be necessary to incorporate further model degrees of freedom (DoF). To study human movements it is therefore often necessary to create models of the human body that are specially suited to the motion or the specific subject being considered. 

There exist several examples of software frameworks that allow for the analysis of movements using rigid-body models (e.g. RBDL \cite{Felis2016}, Puppeteer \cite{Felis2015}, Simbody \cite{ShermanSimbody}, Simox \cite{VahrenkampSimox} and the HuMAns Toolbox \cite{wieberHumans}). However, the creation of models for usage with such softwares is left up to the user. Model creation often involves tedious, manual editing of XML or other formatted files to setup the model kinematics and inertial properties. While this approach may be feasible for a single-use model, for example that of a robot or generic-human that will not be changed frequently, it is unrealistic for modeling large numbers of body types, joint variations and model configurations. For example, an application that would require many models could be a study that involves model-based analysis of experimental data recorded from many subjects (e.g. \cite{Harant2017}). In order to accurately analyze the recorded motions, we need to create subject-specific models corresponding to the body shape and proportions of each participant. Another application could be creating model-variations that differ slightly but systematically from each other. For example, that of a human and a human-model variation including a prosthesis \cite{Kleesattel2017}. To the best of our knowledge, there exists a gap in software tools that allow users to efficiently create large numbers and variations of subject-specific multi-body human models. 

Here, we detail such a model-creation toolbox consisting of scripts written in Matlab/Octave. The collection of scripts, named ModelFactory, allow for a modular and flexible manner to define model configurations, choose from an available set of rules to compute segment properties, and visualize and export the corresponding models. The toolbox also provides functionality to adjust the model kinematics and inertial properties to subject-specific characteristics.

\begin{figure}[ht]
\centering
\includegraphics[width=0.95\textwidth]{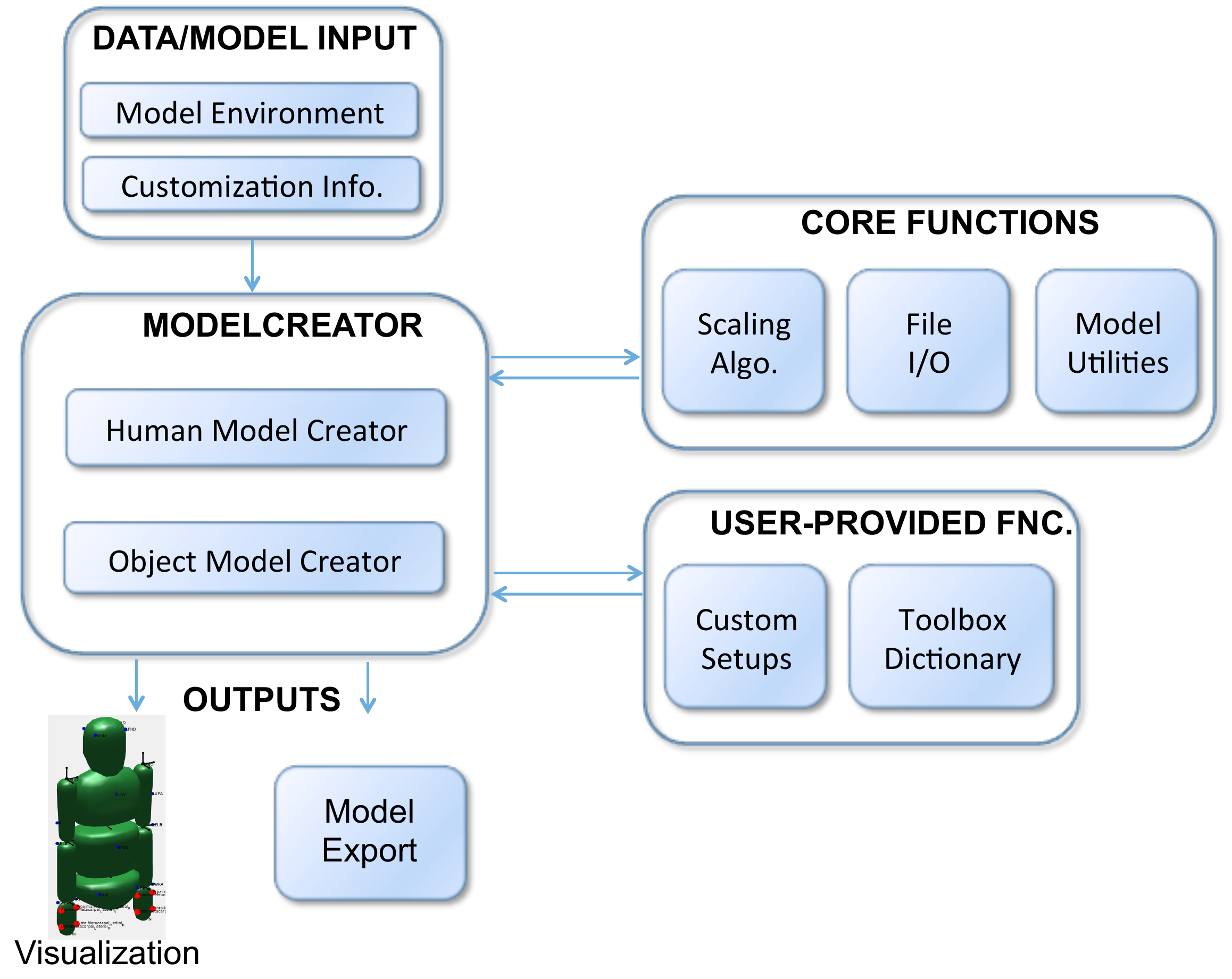}
\caption{ModelFactory overview: The toolbox components can be divided into core functions related to model setup, file input/output and other utilities; user-provided functions such as custom setups; and the information pertaining to a specific model. The ModelCreator script communicates with these components to generate the model file for visualization and export.} 
\label{fig:overview}
\end{figure}

\section{Implementation}
\label{sec:Implementation}
The main part of the toolbox, the ModelCreator, receives input from several components which contributes towards generating the final model (Fig. \ref{fig:overview}). These components provide mandatory information (e.g. results from scaling algorithm, subject anthropometry), and/or optional information that may be used to further refine the model (e.g. custom segment lengths). In the following, we describe the toolbox components in further detail.
\subsection{Dictionary}
The basic building blocks for creating a model are available as predefined ``descriptors" of commonly used joints, points and constraints. This built-in dictionary facilitates the easy setup of a large variety of models by choosing the right combination of descriptors. The following descriptor types are available as default:
\subsubsection*{Joint Types}
Depending on the movement and body segment under consideration, various combinations of rotational and translational joints may be needed. Individual joint types for each segment may be specified by using the corresponding descriptors. For example the descriptor ``Joint\_RY" describes a 1 DoF rotational joint about the Y-axis which is defined in the dictionary by the vector, $[0 \hskip0.5em 1 \hskip0.5em 0 \hskip0.5em 0 \hskip0.5em 0 \hskip0.5em 0]$. The first 3 indices of this vector indicate the rotational axes X, Y and Z, and the last three indicate translational axes X, Y and Z. Similarly, ``Joint\_Root2D\_TXTZRY" describes a 3 DoF joint with translational axes along X and Z and rotation about Y. This joint type may be used to describe the floating base joint of a planar model, and is defined as: 
\begin{equation}
\begin{bmatrix}
    0 & 0 & 0 & 1 & 0 & 0 \\
    0 & 0 & 0 & 0 & 0 & 1 \\
    0 & 1 & 0 & 0 & 0 & 0
\end{bmatrix}
\label{eq:jointTypes}
\end{equation}
with each row of the matrix above corresponding to one of the DoFs. Note that here we follow the spatial vector formulation of Featherstone \cite{FeatherstoneBook}, which is also used in the multi-body dynamics software RBDL \cite{Felis2016}. Commonly used joint types are predefined in the dictionary file \emph{customSetups/dict/dict\_joint\_sets.m} (see toolbox-folder structure in Fig. \ref{fig:folderstructure}).
\subsubsection*{Points}
Pre-defined descriptors allow a user to setup typically used points for each segment. Here, segments are rigid-bodies associated to a local coordinate frame (the description of body segments is detailed later in Sec. \ref{subsec:modelDescriptionFile}). For example, the descriptor ``Points\_Hand\_R\_3D" defines the following points on the right hand segment:

\begin{equation}
\begin{bmatrix}
    ProximalMetacarpal\_Medial\_R & \{-0.2, 0.15, -0.2\}\\
    ProximalMetacarpal\_Lateral\_R & \{0.2, 0.15, -0.2\}\\
	DistalMetacarpal\_Medial\_R & \{-0.2, 0.15, -0.6\}\\
	DistalMetacarpal\_Lateral\_R & \{0.2, 0.15, -0.6\}
\end{bmatrix}
\label{eq:points}
\end{equation}

where, the vector corresponding to each point defines the position of that point in the local segment coordinate frame, when scaled by the segment length. For the foot segment, we allow further detail to be provided that takes subject anthropometry into account (e.g. recorded values for heel-ankle-offset, foot-width etc from the provided  anthropometry). Implementation details of foot points are available in the toolbox documentation. Commonly used points are predefined in the dictionary file \emph{customSetups/dict/dict\_point\_sets.m}.
\subsubsection*{Point Constraints}
\label{subsec:pointConstraints}
Points on a body (human or object) can be constrained to the environment by defining constraint-sets that specify the point name, and the normal along which the constraint should act in base coordinates. For example, in order to constrain a sagittal plane foot segment containing the points ``Toe\_Sagittal" and ``Heel\_Sagittal", we may use the descriptor ``ConstraintSet\_Foot\_Sagittal" which defines the following constraints:

\begin{equation}
\begin{bmatrix}
    FootFlat\_Sagittal & \begin{bmatrix}
 							Heel\_Sagittal & \{1,0,0\} \\
							Heel\_Sagittal & \{0,0,1\} \\
							Toe\_Sagittal  & \{0,0,1\} \\				
    \end{bmatrix} \\
    & \\
    HeelFixed\_Sagittal & \begin{bmatrix}
 							Heel\_Sagittal & \{1,0,0\} \\
							Heel\_Sagittal & \{0,0,1\} \\
    \end{bmatrix} \\
    & \\
    ToeFixed\_Sagittal & \begin{bmatrix}
 							Toe\_Sagittal & \{1,0,0\} \\
							Toe\_Sagittal & \{0,0,1\} \\
    \end{bmatrix}
\end{bmatrix}
\label{eq:pointConstraints}
\end{equation}

where the subset ``FootFlat\_Sagittal" constrains the ``Heel\_Sagittal" point along the X and Z normal directions, and as well the ``Toe\_Sagittal" point along the Z normal direction. This constraint-subset may be used to rigidly fix the foot to the ground at these points. Similarly, rotation of the foot about the Heel\_Sagittal point may be expressed by the ``HeelFixed\_Sagittal" constraint. In this manner, the constraint-set ``ConstraintSet\_Foot\_Sagittal" may be used in a multi-body simulation to describe the behavior of the foot during walking as a series of heel and toe contacts (e.g. \cite{Sreenivasa2017}). Commonly used point constraints are predefined in the dictionary file \emph{customSetups/dict/dict\_constraint\_sets.m}. 

\begin{figure}[ht]
\centering
\includegraphics[width=0.5\textwidth]{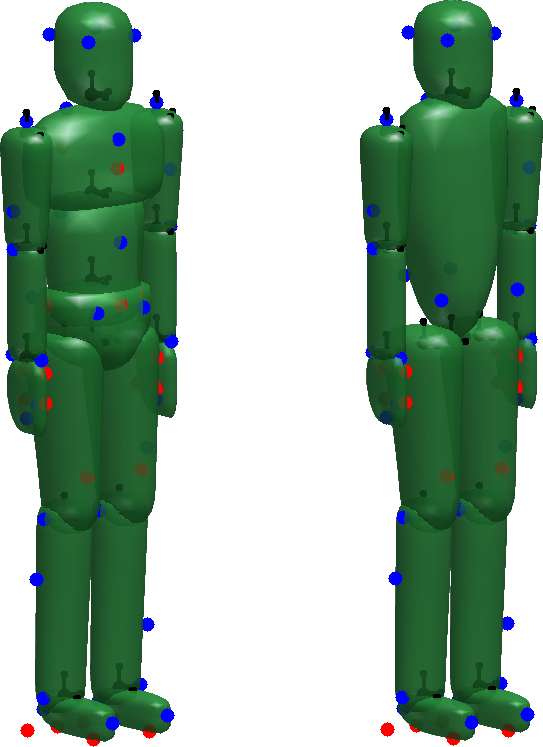}
\caption{Human scaling algorithms: [Left] Model with torso segmented into upper trunk, middle trunk and pelvis segments. [Right] Model with fused torso. Models shown are derived from the equations specified by de Leva 1996 \protect\cite{DeLeva1996}. Blue circles indicate motion capture markers and red circles indicate points on segments. Also shown are the local coordinate frames of the segments.}
\label{fig:scalingAlgos}
\end{figure}

\subsection{Model Types}
Two types of models may be defined; Human and Object. Each of these models may consist of a user-defined number of segments, and each segment derives its common, base properties from the class ``class\_modelSegment" (defined in \emph{core/classes/class\_modelSegment.m}). The properties of this class include:
\begin{itemize} 
\item
segment name, segment parent name, segment parent ID
\item
segment mass, com and inertia properties
\item
segment mesh details for visualization
\item
marker names and positions
\item
joint type, joint center and axis (relative to parent)
\item
segment length (used for scaling purposes in human model)
\item
associated points
\item
associated constraint sets
\end{itemize} 
\subsubsection{Human model} 
\label{subsubsec:humanmodel}
We define scaling algorithms that provide methods to compute the segment kinematics and inertial properties from a person's anthropometric details. Examples of scaling algorithm are the linear regression equations from de Leva \cite{DeLeva1996}, Dempster \cite{Dempster1955}, Jensen \cite{Jensen1986} and Zatsiorsky \cite{Zatsiorsky1983}. Currently, the scaling algorithms from de Leva and Jensen have been made available as toolbox options. The de Leva scaling provides nominal values for segment lengths, and mass and inertia properties based on subject gender, height and weight. We provide 3 de Leva datasets, the first two of which treat the torso as either one segment or subdivided into 3 segments (pelvis, mid-trunk, upper-trunk), Fig. \ref{fig:scalingAlgos}. A third de Leva derived dataset can be used to define models in the sagittal plane where the inertial properties of the left and right limbs are combined. The scaling equations by Jensen are better suited for modeling the body segment properties of children. Jensen's equation take into account the child's age and weight, but requires the segment lengths to be provided by the user. Additional scaling algorithms can be defined by the user using the templates provided with the toolbox. Note that all user-defined functions should be made available to the toolbox in the \emph{customSetups} folder.

In order to use the provided scaling algorithms, the user needs to provide some anthropometric details of the person being modeled. General details such as age, height, weight and gender are then used to compute the nominal proportions of the human body segments in the sagittal plane. Further information about the human, such as pelvis width, distance between hip centers, distance between shoulder centers, and foot offsets may be used to setup specialized joint centers for the shoulders, hips and feet. Anthropometric details are specified in a file with keywords as described below:

\begingroup
\obeylines
\begin{Verbatim}
# Data specifying anthropometric details of subject
# Age in years
age, 30.0
# Height in m
height, 1.7
# Weight in kg
weight, 80.0
# Gender (1 = Male, 0 = Female)
gender, 1
# Maximum width of pelvis in m
pelvisWidth, 0.24
# Distance between hip centers in m
hipCenterWidth, 0.17
# Distance between shoulder centers in m
shoulderCenterWidth, 0.4
# Distance between heel and ankle center along foot length in m
heelAnkleXOffset, 0.08
# Vertical distance between heel and ankle center in m
heelAnkleZOffset, 0.09
# Vertical distance between shoulder centers and C7 in m
shoulderNeckZOffset, 0.07
# Maximum width of foot in m
footWidth, 0.09
\end{Verbatim}
\endgroup

Note that some of the fields in the anthropometry file may not be required for certain model types. For example, if a custom scaling algorithm is defined that can compute hip and shoulder widths, then these may be omitted from the anthropometry file. Similarly, all values pertinent to the transverse plane are unnecessary if one is defining a planar model in the sagittal plane. It is left upto the user to choose the right combination of anthropometric detail and scaling algorithm, with the toolbox providing corresponding checks and error messages in case of missing information.
\subsubsection{Object model}
Object models can be used to define any number of bodies such as boxes, orthoses, or exoskeletons that may be used in conjunction with the human model. Each object is associated with a setup function that defines the configuration of the individual segments that make up the object. This setup function is analogous to the scaling algorithm used to specify the details of the human body segments. Segment inertial properties for objects may be defined in several ways. First, they may be directly included in the setup function. Second, they may be computed automatically using a mean density per segment and the segment mesh volume. Third, they may be provided directly by the user while defining a specific object. This flexibility in computing inertial properties allows the user to define a wide range of objects and properties. 

Inertial properties based on mean-density or user-values can be specified in a file as follows:

\begin{center}
\begin{tabular}{|lllll|}
\hline
Segment1, & UseMeanDensity, & density, & ,,, & ,,,  \\
Segment2, & UseUserValues, & mass, & CoM, & inertia \\ \hline
\end{tabular}
\end{center}

If the keyword ``UseMeanDensity" is specified, the mean density value in $kg/m^3$ and the mesh volume of the segment is used to compute mass, center of mass and inertia. If ``UseUserValues" is specified then the values entered in the file are used, where mass is a positive scalar in kg, CoM is a 3-dimensional vector specifying the center of mass of the segment in the local coordinates, and inertia is a 9-dimensional vector that specifies each row of the inertia matrix. Examples of object mass properties files are available in \emph{data/samples/3DHumanExoBox/ModelFiles...}.

\begin{figure}[h!]
\centering
\includegraphics[width=0.5\textwidth]{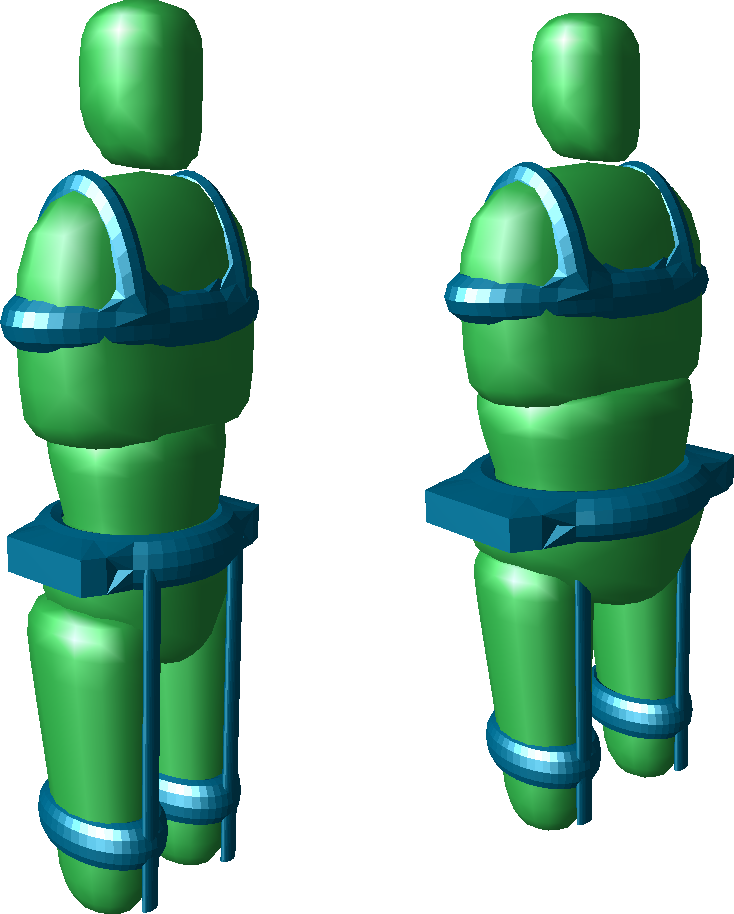}
\caption{Exoskeletons scaled to the human model may be created using the corresponding functions (samples under \emph{customSetups/setups/exo}). Shown is the automatic scaling of a 3D exoskeleton to two human models with varying segment proportions.}
\label{fig:exoModels}
\end{figure}

Some examples of object setups are included that specify exoskeletons and box-type weights. For example, setups to create exoskeleton models that correspond to a sagittal plane human model, as well as those for a 3D human model (Fig. \ref{fig:exoModels}). Both exoskeleton setups automatically scale the exoskeleton to the human model. Sample setups for creating box objects are provided in the toolbox under \emph{customSetups/setups/box} (sample object shown in Fig. \ref{fig:gui}).

Note that currently we only model the kinematics and inertial properties of the objects, and not more detailed aspects such as spring-loaded joints and actuator characteristics (e.g. for exoskeletons). These model enhancements will be the focus of future developments of the ModelFactory toolbox.
\subsection{Model Description}
\label{subsec:modelDescriptionFile}
Both human and object models are described in a file that details the kinematic chain, the number and type of segments, joint types etc. Other than the segment names, all descriptors must correspond to elements in the dictionary, the scaling algorithm (in case of the human model) or the object setup (in case of the object model). An example of the description file for a human model is shown below. 

\begingroup
\obeylines
\begin{Verbatim}
Pelvis, Segment_Pelvis, Joint_Root3D_TXTYTZRYRXRZ, ROOT, Points_Pelvis_3D, ,
Thigh_R, Segment_Thigh_R, Joint_RYRXRZ, Pelvis, Points_Thigh_R_3D, ,
Shank_R, Segment_Shank_R, Joint_RY, Thigh_R, , ,
Foot_R, Segment_Foot_R, Joint_RYRX, Shank_R, Points_Foot_R_3D, ConstraintSet_Foot_R_3D,
Thigh_L, Segment_Thigh_L, Joint_RYRXRZ, Pelvis, Points_Thigh_L_3D, ,
Shank_L, Segment_Shank_L, Joint_RY, Thigh_L, , ,
Foot_L, Segment_Foot_L, Joint_RYRX, Shank_L, Points_Foot_L_3D, ConstraintSet_Foot_L_3D,
MiddleTrunk, Segment_MiddleTrunk, Joint_RYRXRZ, Pelvis, , ,
UpperTrunk, Segment_UpperTrunk, Joint_RYRXRZ, MiddleTrunk, Points_UpperTrunk_3D, ,
Head, Segment_Head, Joint_RYRXRZ, UpperTrunk, , ,
UpperArm_R, Segment_UpperArm_R, Joint_RYRXRZ, UpperTrunk, , ,
LowerArm_R, Segment_LowerArm_R, Joint_RYRZ, UpperArm_R, , , 
Hand_R, Segment_Hand_R, Joint_RYRXRZ, LowerArm_R, Points_Hand_R_3D, ConstraintSet_Hand_R_3D,
UpperArm_L, Segment_UpperArm_L, Joint_RYRXRZ, UpperTrunk, , ,
LowerArm_L, Segment_LowerArm_L, Joint_RYRZ, UpperArm_L, , ,
Hand_L, Segment_Hand_L, Joint_RYRXRZ, LowerArm_L, Points_Hand_L_3D, ConstraintSet_Hand_L_3D,
\end{Verbatim}
\endgroup

Each line has the following fixed structure:
\begin{enumerate}
 \item Name of the segment (may be freely chosen)
 \item Segment type as it is defined in the scaling algorithms or the setup functions for objects
 \item Joint type that is connected to the segment. 'R' denotes a rotational joint and 'T' a translation joint. 'X, Y and Z' are the frontal, transverse and longitudinal axis respectively
 \item Name of the parent segment. The predefined name for the origin is ROOT
 \item Set of points (optional). The points have to be defined in the dictionary ``dict\_point\_sets.m" or made available as custom point definitions
 \item Set of constraints (optional). The constraints have to be included in the dictionary ``dict\_constraint\_sets.m" or made available as custom constraint definitions
\end{enumerate}

\subsection{Model Customization}
We allow for the customization of model proportions to better match specific individuals. In addition, we allow for definition of custom marker setups and dictionary descriptors.
\subsubsection{Custom scaling of human model segments}
\begin{figure}[ht]
\centering
\includegraphics[width=0.5\textwidth]{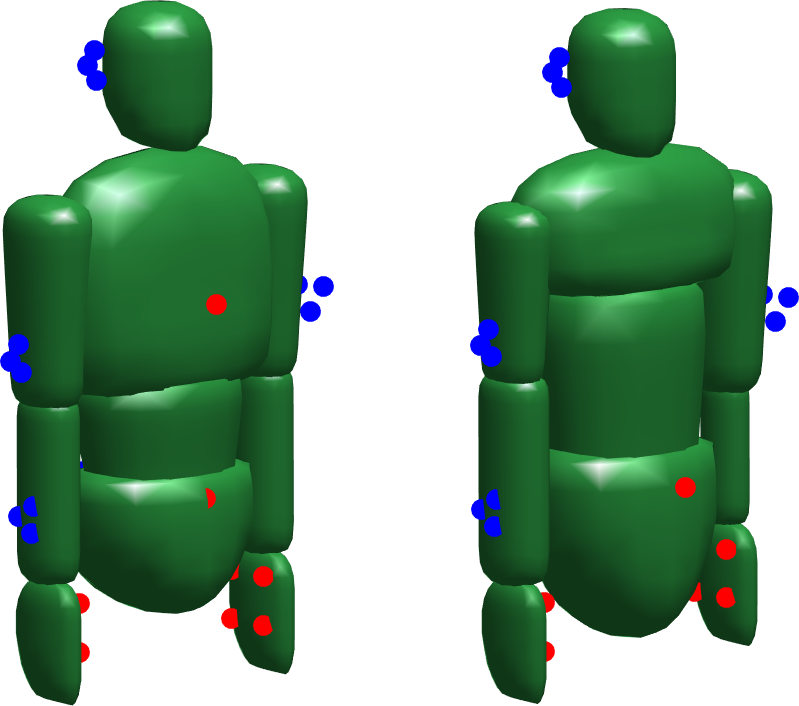}
\caption{Custom scaling may be used to change the default proportions of segment lengths [Left], or to create subject-specific and asymmetric models [Right]. Also shown is a custom marker setup (blue circles).}
\label{fig:customScaling}
\end{figure}

Human model segments may be scaled to subject-specific values by providing a list of segment lengths and the corresponding segment names. These custom segment lengths, $lcustom$, may for example be measured in an experimental setting. Additional customization values in the transverse plane (e.g hip width and shoulder width) can be provided as part of the subject anthropometry. ModelCreator updates the model kinematics to reflect the provided custom scaling. In addition, the segment mass distributions are proportionally adjusted to match the relative segment lengths. This is done such that the adjusted segment masses sum up to the total body mass provided in the subject anthropometry, as follows: 
\begin{equation}
mcustom_i = mdefault_i \left( \frac{lcustom_i}{ldefault_i} \right)\left( \frac{M}{M_{unadj}} \right), \qquad i = 1,...,N
\end{equation}
where, $N$ is the number of segments, $mcustom_i$ and $mdefault_i$ denote the custom and default segment masses. $lcustom_i$ and $ldefault_i$ are the corresponding custom and default segment lengths. $M$ denotes the total mass of the human, and $M_{unadj}$ is computed as:
\begin{equation}
M_{unadj} = \sum_{i=1}^N mdefault_i \left( \frac{lcustom_i}{ldefault_i} \right)
\end{equation}
Figure \ref{fig:customScaling} shows some examples of customized models with different segment lengths and proportions. Segment lengths are specified as a formatted text file containing the segment length and name, as shown below:

\begingroup
\obeylines
\begin{Verbatim}
# The order of segment names here does not matter.
0.15, Pelvis
0.40, Thigh_R
0.40, Shank_R
0.29, Foot_R
0.40, Thigh_L
0.40, Shank_L
0.29, Foot_L
0.14, MiddleTrunk
0.34, UpperTrunk
0.20, Head
0.286, UpperArm_R
0.256, LowerArm_R
0.18, Hand_R
0.286, UpperArm_L
0.256, LowerArm_L
0.18, Hand_L
\end{Verbatim}
\endgroup

\subsubsection{Motion-capture markersets}
Motion-capture markers may be used to reconstruct the model motion from recorded data with methods such as inverse kinematics based on least-squares optimization (e.g. \cite{Felis2015}). Marker definitions corresponding to the VICON PIG markerset \cite{ViconPIG} may be included in the human model (Fig. \ref{fig:scalingAlgos}). This option is controlled via the ``Addmarkers" field while setting up the model (detailed later in Sec. \ref{subsec:envFile}). Alternatively, customized configurations of markers may be used by defining the marker placement on each of the human and object segments. This definition is provided as a formatted text file shown in the example below:

\begin{center}
\begin{tabular}{lll}
\hline
\multicolumn{1}{|c}{Segment\_Pelvis,} & Cluster, & 0.043,   \\
\multicolumn{1}{|c}{\emph{name of the segment the}} & \emph{marker type} & \emph{distance between}   \\
\multicolumn{1}{|c}{\emph{marker is attached to}} &  & \emph{markers (for Clusters)}  \\
\hline 
 % & & \\
\hline
  Pelvis\_1, Pelvis\_2, Pelvis\_3, , , , & -1.0, -0.05, 0.90, & \multicolumn{1}{c|}{0, 20, 0}  	\\
\emph{marker names}			& \emph{translational offset} & \multicolumn{1}{c|}{\emph{rotational offset}}	\\
\hline \\
\end{tabular}
\end{center}

The translational and rotational offset is specified with respect to the origin of the body the marker is attached to. This example sets up a customized flange-type marker cluster and is included in the toolbox under \emph{data/samples/ModelFiles} \emph{\_3DHumanCustom/...}.  There are several marker types available: Marker (one single point), Cluster (consisting of three markers), DoubleCluster (consisting of 2 parallel clusters) (refer to Fig. \ref{fig:customScaling}). For further details for specifying custom markers see the toolbox documentation.
\subsubsection{Custom dictionary terms}
The dictionary terms used to create models may be expanded by adding customized definitions of joint types, points, and point constraints using the templates described in Eqs. \eqref{eq:jointTypes}, \eqref{eq:points} and \eqref{eq:pointConstraints}, respectively. User-defined dictionary terms should be made available to the toolbox by editing the file \emph{customSetups/dict/dict\_definitions.m} and adding the location of the file containing the custom terminology.

Custom points may be used to include points on segments that are of interest for specific applications (e.g. bony landmarks or mesh corners for collision checking). Custom joints provide a way for users to use their own terminology for defining joints (i.e. different from the spatial vector formulation \cite{FeatherstoneBook} used as default). The implementation and use of custom joints can be influenced by using the boolean ``customJoint" in the corresponding dictionary descriptor. With this functionality, users can for example, define complex joint movements such as the surface-geometry based scapulo-thoracic joint \cite{Seth2016}. Note that custom joints could possibly require additional model export functions. 

It may also be useful to define custom constraints for use during the computation of multi-body dynamics. Custom constraints may be used by adding the relevant dictionary terms, and defining how they are used in the models and written to the model file. As an example, custom loop constraints have been included in the toolbox. In contrast to the body-to-environment constraints-sets (Sec. \ref{subsec:pointConstraints}), loop constraints may be used to constrain chosen degrees of freedom between two points on two bodies (e.g. human to object or object to object). For usage of point and loop constraints refer to the publications \cite{Sreenivasa2017,Millard2017,Harant2017} and the multi-body dynamics library \cite{Felis2016}.

\begin{figure}[ht]
\centering
\includegraphics[width=0.95\textwidth]{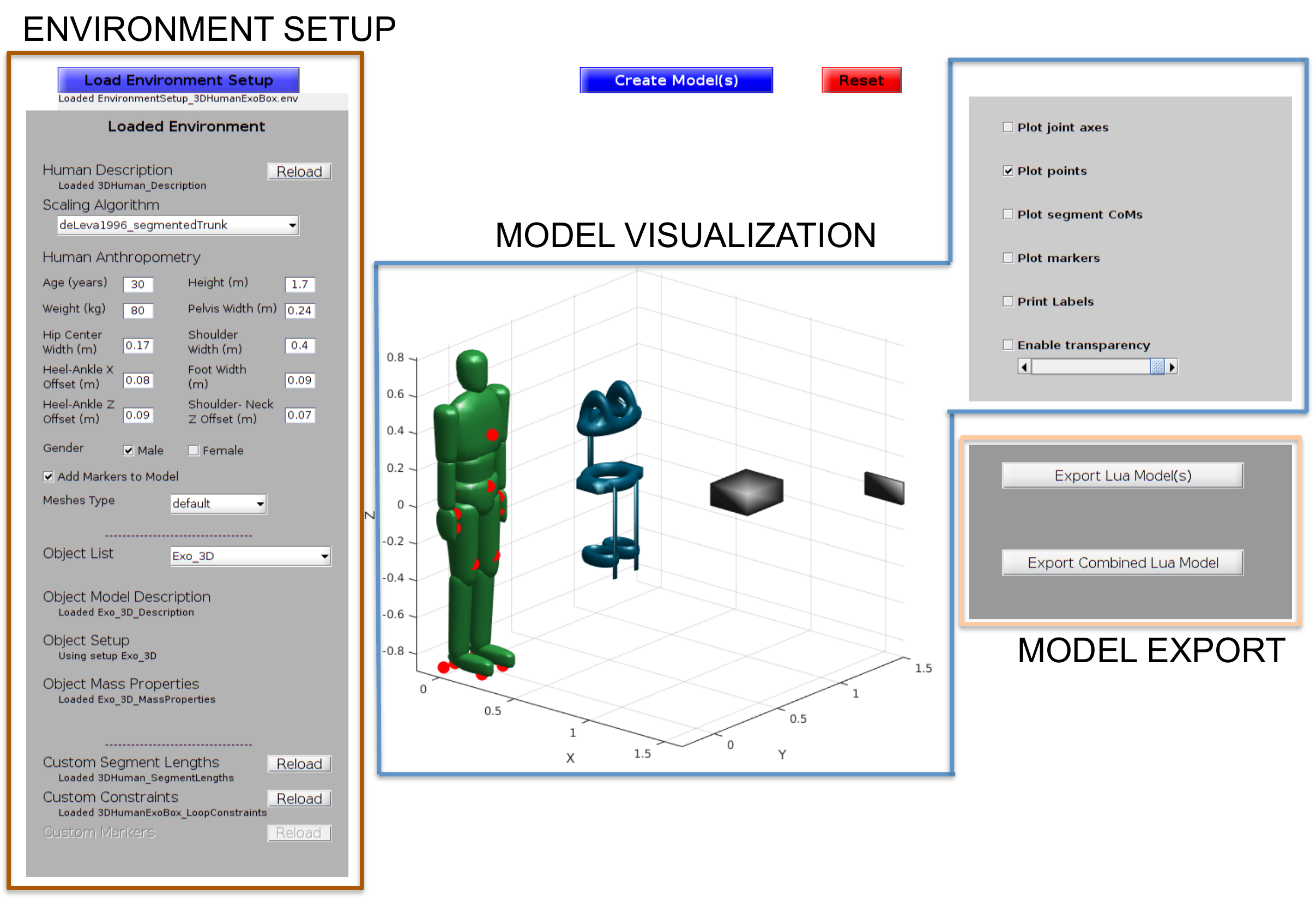} 
\caption{ModelFactory Graphical User Interface under Matlab R2017a - Outlined are the parts related to the environment variables, visualization and model export. The values displayed in the human-related boxes may still be modified before model creation.} 
\label{fig:gui}
\end{figure}

%%%%%%%%%%%%%%%%%%%%%%%%%%%%%%%%%%%%%%%%%%%%%%%%%%%%%%%%%%%%%%%%%%%%%%%%%%%
\section{Results and discussion}
\label{sec:resultsDisc}
The ModelFactory toolbox may be accessed via a graphical user interface (GUI) or by using a text-interface. The GUI is available for usage with Matlab\textregistered via the script ModelCreator\_GUI.m (Fig. \ref{fig:gui}), and the non-GUI version may be used in Matlab and Octave via the script ModelCreator\_noGUI.m. For both GUI and non-GUI versions the same basic set of files are required to setup, create and export models (as described in Sec. \ref{sec:Implementation}). All the model description files and various options are listed in a single environment file as detailed in the following.
\subsection{Environment file} 
\label{subsec:envFile}
The environment file consists of sets of keywords followed by string values associated with the keywords. Each environment file is associated with only one human model and any number of object models. Excerpts from an environment file are provided below followed by details about each of the keywords. We start with keywords related to the human model:
 
\begingroup
\obeylines
\begin{Verbatim}
humanModel_DescriptionFile
./Human_Description

humanModel_ScalingAlgorithmChoice
deLeva1996_segmentedTrunk

humanModel_AnthropometryFile
./Human_Anthropometry

humanModel_UseCustomLengths
./Human_SegmentLengths

humanModel_UseCustomConstraints
./HumanExoBox_LoopConstraints

humanModel_TypeMeshes
human

humanModel_Save
./Human.lua

AddMarkers
1

objectModel1_DescriptionFile
./Exo_Description

objectModel1_SetupChoice
Exo_3D

objectModel1_MassProperties
./Exo_MassProperties

UseCustomMarkers
./HumanExoBox_MarkerOptotrak

combinedModel_Save
./HumanExoBox.lua


\end{Verbatim}
\endgroup

\begin{figure}[ht]
\centering
\includegraphics[width=0.6\textwidth]{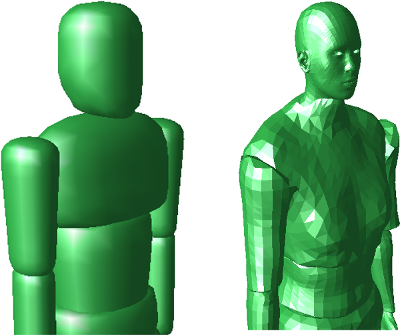}
\caption{Human body meshes as simple geometric elements [Left] or more detailed meshes derived from the MakeHuman \protect\cite{MakeHuman} software [Right]}.
\label{fig:humanMeshes}
\end{figure}

Most of the keywords are self explanatory and refer to the human model aspects detailed in Sec. \ref{sec:Implementation}. With the field ``humanModel\_TypeMeshes" one can choose the default geometric shapes or more detailed human-like meshes (Fig. \ref{fig:humanMeshes}) derived from the software MakeHuman \cite{MakeHuman} to visualize the human model. The field ``AddMarkers" adds the default markerset to the human model. The object model keywords are as follows:

\begingroup
\obeylines
\begin{Verbatim}
humanModel_DescriptionFile
./Human_Description

humanModel_ScalingAlgorithmChoice
deLeva1996_segmentedTrunk

humanModel_AnthropometryFile
./Human_Anthropometry

humanModel_UseCustomLengths
./Human_SegmentLengths

humanModel_UseCustomConstraints
./HumanExoBox_LoopConstraints

humanModel_TypeMeshes
human

humanModel_Save
./Human.lua

AddMarkers
1

objectModel1_DescriptionFile
./Exo_Description

objectModel1_SetupChoice
Exo_3D

objectModel1_MassProperties
./Exo_MassProperties

UseCustomMarkers
./HumanExoBox_MarkerOptotrak

combinedModel_Save
./HumanExoBox.lua


\end{Verbatim}
\endgroup

Consecutive objects can be defined by numbering the related object files. Additionally, the following general environment keywords may be defined:

\begingroup
\obeylines
\begin{Verbatim}
humanModel_DescriptionFile
./Human_Description

humanModel_ScalingAlgorithmChoice
deLeva1996_segmentedTrunk

humanModel_AnthropometryFile
./Human_Anthropometry

humanModel_UseCustomLengths
./Human_SegmentLengths

humanModel_UseCustomConstraints
./HumanExoBox_LoopConstraints

humanModel_TypeMeshes
human

humanModel_Save
./Human.lua

AddMarkers
1

objectModel1_DescriptionFile
./Exo_Description

objectModel1_SetupChoice
Exo_3D

objectModel1_MassProperties
./Exo_MassProperties

UseCustomMarkers
./HumanExoBox_MarkerOptotrak

combinedModel_Save
./HumanExoBox.lua


\end{Verbatim}
\endgroup

where, a custom markerset may be used by providing the path to the marker definition file with the field ``UseCustomMarkers". The keyword ``combinedModel\_Save" collates all the models (human + all objects) into one Lua file.

\subsection{Model creation and export}
\label{subsec:modelCreationExport}
Environment files are loaded by ModelCreator and the options are processed to create the corresponding models. The human and object models are created sequentially and separately. Note that the definition and creation of objects is optional, and the only mandatory fields to create the human model are the anthropometric information (Sec. \ref{subsubsec:humanmodel}), human model description file (Sec. \ref{subsec:modelDescriptionFile}) and the choice of scaling algorithm (Sec. \ref{subsubsec:humanmodel}).

\begin{figure}[ht]
\centering
\includegraphics[width=0.90\textwidth]{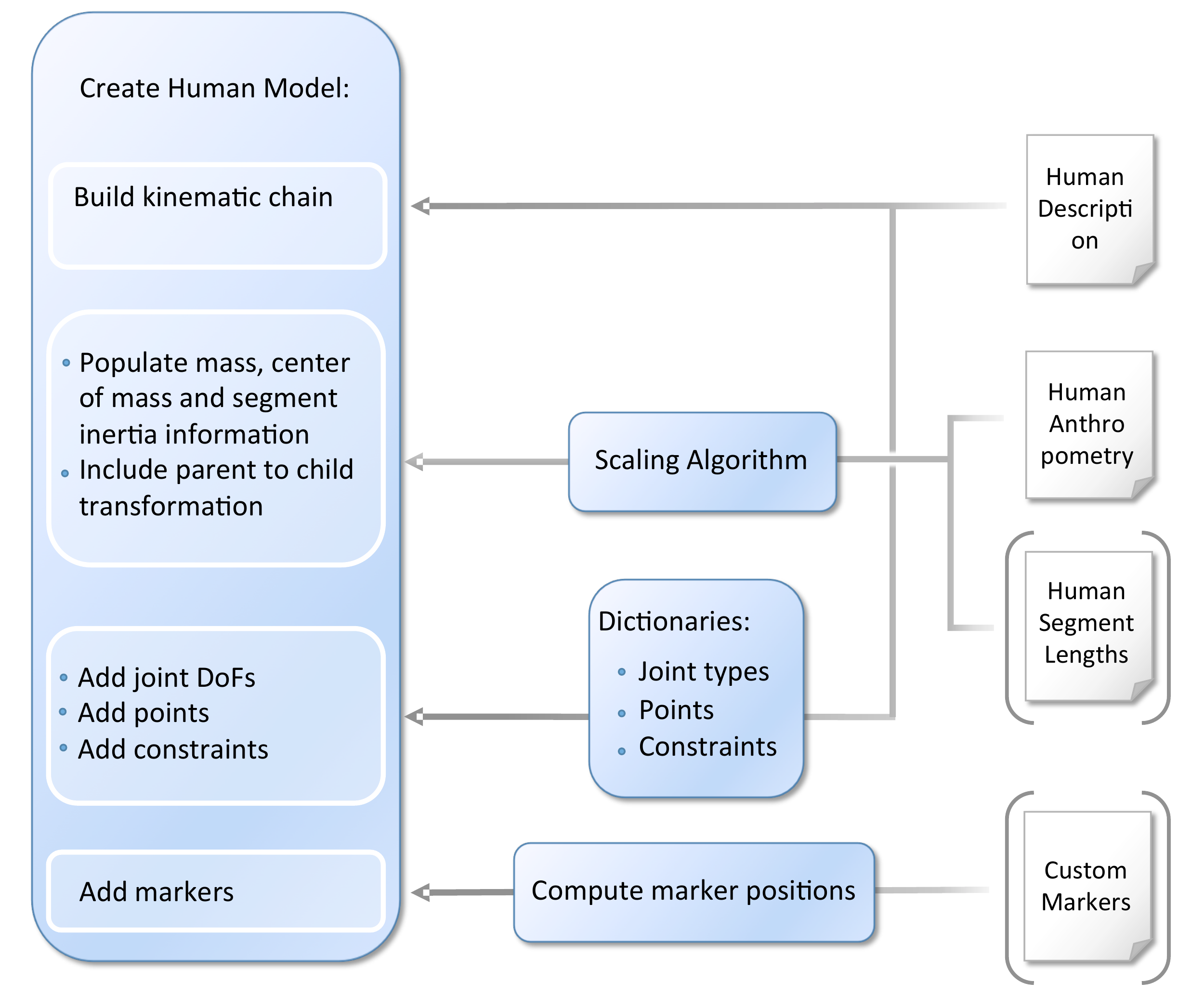}
\caption{General procedure for creating human models. Mandatory fields are the anthropometry, model description and choice of scaling algorithm. Object model creation follows a similar procedure.} \label{fig:modelCreation}
\end{figure}

The general procedure for human model creation is shown in Fig. \ref{fig:modelCreation}. Objects are created similarly but with fewer available functionalities as detailed in Tab. \ref{tab:Functionalities}. First, we build the kinematic chain of the model based on the description file. The individual segments then get populated with data from a scaling algorithm (or object setup), the transformations between parent and child body, the joint DoFs connecting the parent and child, the points (if any) and the point constraints (if any). Finally, if defined, the position of the markers are computed and added to the model. 

After successful model creation, the results may be exported into individual Lua files or combined together in one Lua file. These options are controlled via the buttons provided in the GUI (Fig. \ref{fig:gui}) or via the ``humanModel\_Save" and ``combinedModel\_Save" keywords in the environment file. 
\subsection{Use case}
\label{subsec:UseCase}
\begin{figure}[h]
\centering
\includegraphics[width=0.5\textwidth]{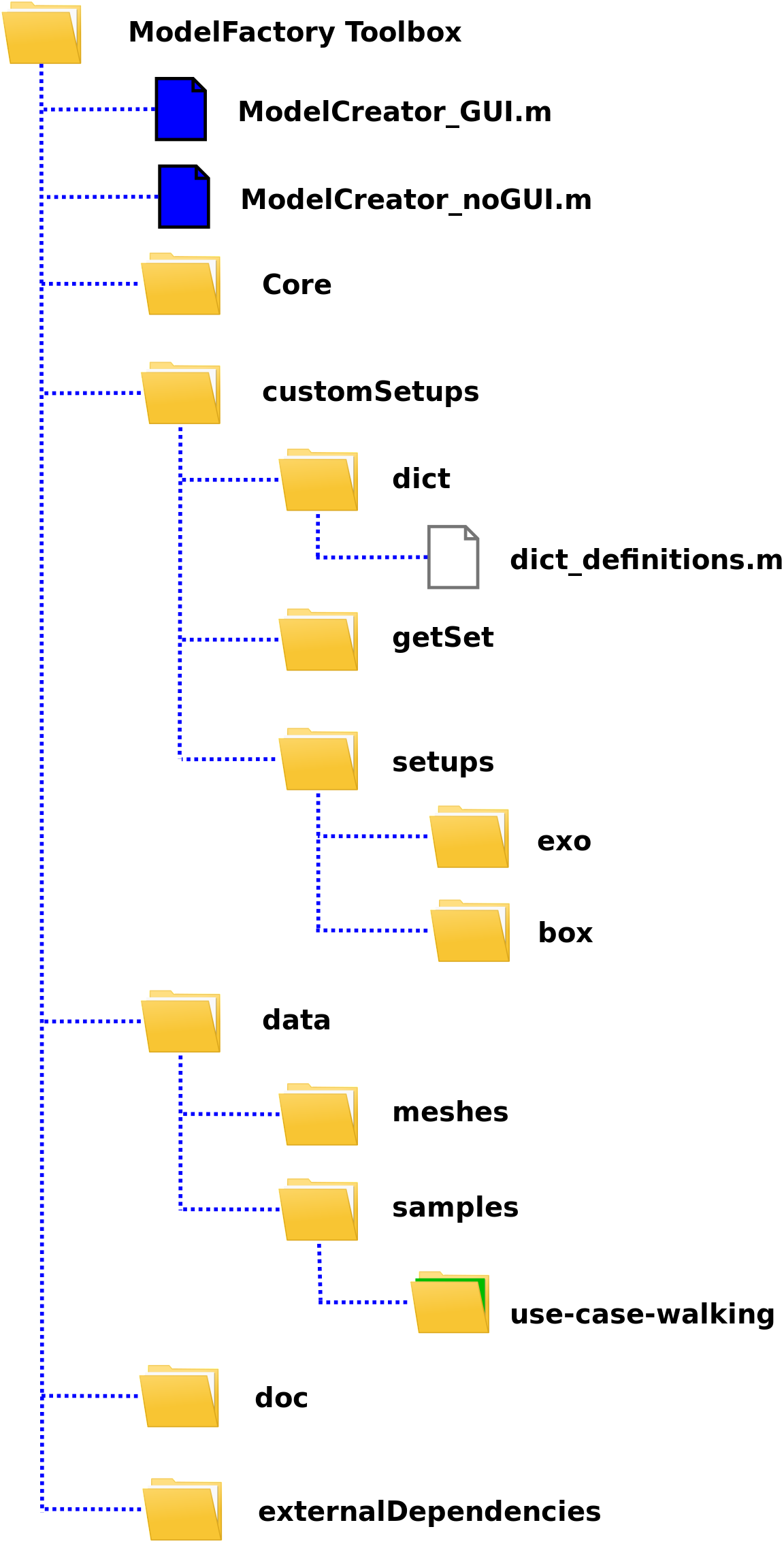}
\caption{Overview of the folder structure of the ModelFactory toolbox and location of important files. Most user-related files are present in the \emph{customSetups} folder. For example, new dictionary definitions should be added to the \emph{dict\_definition.m} script, and new object setup functions should be added to the \emph{customSetups/setups} folder. Several sample environment files and a use-case are available in the \emph{data/samples} folder. The meshes used to visualize the model are available in the \emph{data/meshes} folder.}
\label{fig:folderstructure}
\end{figure}

The toolbox includes a wide range of sample environment and model files under the folder \emph{data/samples}, as well as additional documentation (see toolbox-folder structure in Fig. \ref{fig:folderstructure}). We also include an example where a human model and experimentally recorded walking data is used to compute joint angle and joint torques using inverse kinematics and inverse dynamics. Inverse kinematics is solved as a least-squares optimization problem using the open-source software Puppeteer\footnote{https://github.com/martinfelis/puppeteer} \cite{Felis2015}. Inverse dynamics is computed using the open-source rigid-body dynamics library RBDL\footnote{https://bitbucket.org/rbdl/rbdl/} \cite{Felis2016}. Both Puppeteer and RBDL are compatible with the Lua model export format currently offered by the ModelFactory toolbox. Additional code is included for reading the model fields such as points and constraints that were described in previous sections. This use-case example is available in the folder \emph{data/samples/use-case-walking}. 

\subsection{Future Developments}
\label{subsec:futureDev}
Further developments of the toolbox are planned in several ways. First, it would be interesting to include other model export formats (e.g URDF) to allow the models to be more widely used. The toolbox functionality could be extended to include model fields related to actuation. For humans this could be joint actuators such as muscle torque generators \cite{Millard2017} and line-type muscle models \cite{Delp2007}. In this context, the models of objects with actuators (e.g. active exoskeletons, prostheses) could also be interesting. This would open up the  possibility of creating robot models alongside their motor characteristics, sensors etc. The model applications considered here are limited to the kinematics and dynamics of rigid-body systems. In general, it is of interest to incorporate deformable models and those with wobbling masses (e.g. \cite{Gruber1998}). Some of the developement ideas mentioned here are the focus of ongoing work and will be published on the public repository as available.
%
%%%%%%%%%%%%%%%%%%%%%%%%%%%%%%%%%%%%%%%%%%%%%%%%%%%%%%%%%%%%%%%%%%%%%%%%%%%
\section{Conclusions}
\label{sec:conclusions}
The ModelFactory toolbox detailed here allows users to create a wide range of multi-body models in a quick and standardized manner. For applications requiring a number of subject-specific human models (e.g. \cite{Harant2017}), or those where a model is varied in a systematic manner, such a model creation toolbox can save the user time and effort. Using the extensive examples as templates, users can also extend the toolbox to include model customizations specific to their application. Note that the ModelFactory toolbox only generates model files that may be used to conduct further analysis of human movements. For example, with methods such as inverse kinematics and inverse dynamics (e.g. \cite{Sreenivasa2016}). Alternatively the models can be used to simulate movements with methods such as optimal control \cite{Sreenivasa2017,Millard2017}. The toolbox is published as open-source software and is compatible with other open-source softwares (e.g. RBDL \cite{Felis2016} and Puppeteer \cite{Felis2015}) that can be used with the generated model files.

\section{Software availability}
Repository: https://github.com/manishsreenivasa/ModelFactory

Operating system(s): Platform independent

Programming language: Matlab / Octave

License: zLib (\url{http://www.zlib.net/zlib_license.html})

\section{List of abbreviations}
{\bf GUI} - Graphical User Interface; {\bf RBDL} - Rigid Body Dynamics Library; {\bf URDF} - Unified Robot Description Format; {\bf XML} - Extensible Markup Language; {\bf DoF} - Degree of Freedom

\renewcommand{\arraystretch}{1.5}
\begin{table}[htb]
\caption{Overview of the functionalities currently provided in the toolbox for the two model types. 'X' marks the availability of the functionality for that model type.} \label{tab:Functionalities}
\begin{center}
\begin{tabular}{|l|c|c|}
\hline
& \multicolumn{2}{c|}{\emph{Model Type}} \\ \cline{2-3}
\emph{Functionality} 	& Human & Object \\ \hline
Anthropometry 			& X 	& 	\\ \hline
Model description 		& X		& X	\\ \hline
Scaling algorithms 		& X		& 	\\ \hline
Custom scaling 			& X		& 	\\ \hline
Joint types 			& X	 	& X	\\ \hline
Points		 			& X 	& X	\\ \hline
Point constraints 		& X		& X	\\ \hline
Custom markers 			& X		& X	\\ \hline
Custom setups 			& 		& X	\\ \hline
Segment mass from mesh 	& 		& X	\\ \hline
Segment mass from user 	& 		& X	\\ \hline
\end{tabular}
\end{center}
\end{table}
\renewcommand{\arraystretch}{1.0}

\subsection*{Availability of data and material}
The data relevant to this work are included in this article's supplementary information files. Future developments are available on the ModelFactory repository, \url{https://github.com/manishsreenivasa/ModelFactory}.
\subsection*{Competing interests}
The authors declare that they have no competing interests.
\subsection*{Funding}
Financial support by the European Commission within the H2020 project SPEXOR (GA 687662) is gratefully acknowledged.
\subsection*{Authors' contributions}
MS developed the software architecture with assistance from MH. Both authors read and approved the final manuscript.
\subsection*{Acknowledgements}
We thank Matthew Millard, Kevin Stein and Katja Mombaur for assistance with software development, contributions to the code and helpful discussions.

\bibliographystyle{ieeetr}
\bibliography{modelFactoryBib}

\end{document}